\documentclass[10pt,letterpaper,conference]{IEEEtran}
\usepackage[utf8]{inputenc}
\usepackage{%
    amsmath,amssymb,amsthm,mathtools,stmaryrd,
    comment,xspace,hyperref,cleveref,
    algorithm,algorithmicx,algpseudocode,
    placeins,
}
\usepackage{pdflscape}
\usepackage{verbatim}
\usepackage{pgfplots}
\DeclareUnicodeCharacter{2212}{−}
\usepgfplotslibrary{groupplots,dateplot}
\usetikzlibrary{patterns,shapes.arrows}
\pgfplotsset{compat=newest}

\usepackage[textsize=scriptsize
           ]{todonotes}

\title{Bayesian Optimisation with Formal Guarantees}
\author{%
     \IEEEauthorblockN{Franz Brau\ss{}e}
     \IEEEauthorblockA{%
        Department of Computer Science \\
        University of Manchester, UK \\
        franz.brausse@manchester.ac.uk} \and
     \IEEEauthorblockN{Zurab Khasidashvili}
     \IEEEauthorblockA{%
        Product Enablement Solutions Group \\
        Intel Israel Development Center \\
        zurab.khasidashvili@intel.com
     } \and
     \IEEEauthorblockN{Konstantin Korovin}
     \IEEEauthorblockA{%
        Department of Computer Science \\
        University of Manchester, UK \\
        konstantin.korovin@manchester.ac.uk
    }
}
\date{May 21st, 2021}

\newcommand*\dom{\operatorname{dom}}

\theoremstyle{definition}

\newcommand{\todofb}[2][]{\todo[color=cyan!30,tickmarkheight=.2em,#1]{FB: #2}}

\newcommand*\UNSAT{{\small\textsf{unsat}}\xspace}
\newcommand*\BO{{\text{BO}}\xspace}
\newcommand*\Bo{{\text{Bayesian optimization}}\xspace}
\newcommand*\ML{{\text{machine learning}}\xspace}
\newcommand*\no{{\text{near-optimal}}\xspace}
\renewcommand{\vec}[1]{\boldsymbol{#1}}

\newcommand*\Max{^{\max}}
\newcommand*\Min{^{\min}}

\algnewcommand{\algorithmicendif}{\textbf{{f}{i}}}
\algnewcommand{\algorithmicendfor}{\textbf{end}}
\algblockdefx[IF]{If}{EndIf}[1]{\algorithmicif\ #1\ \algorithmicthen}{\algorithmicendif}
\algblockdefx[FOR]{For}{EndFor}[1]{\algorithmicfor\ #1\ \algorithmicdo}{\algorithmicendfor}

\algnewcommand{\IIf}[1]{\State\algorithmicif\ #1\ \algorithmicthen}
\algnewcommand{\EndIIf}{\unskip\ \algorithmicendif}

\begin{document}
\maketitle

\begin{abstract}
    Application domains of Bayesian optimization include optimizing black-box
    functions or very complex functions. The functions we are interested in describe
    complex real-world systems applied in industrial settings. Even though
    they do have explicit representations, standard optimization
    techniques fail to provide validated solutions and correctness
    guarantees for them.
    In this paper we present a combination of Bayesian optimisation and SMT-based constraint solving to achieve safe and stable solutions with optimality guarantees.
\end{abstract}

%%%%%%%%%%%%%%%%%%%%%%%%%%%%%%%%%%%%%%%%%%%%%%%%%%%%%%%%%%%%%%%%%%%%%%%%%%%%%%%
\section{Introduction}\label{sec:intro}
%%%%%%%%%%%%%%%%%%%%%%%%%%%%%%%%%%%%%%%%%%%%%%%%%%%%%%%%%%%%%%%%%%%%%%%%%%%%%%%

$\Bo$ ($\BO$)~\cite{Moc75,Fra18} is a popular technique for optimizing an objective function~$f$, which is done through searching for input points $\vec x$ such that $f(\vec x)$
approximates the maximum $\max f$. We refer to these points $\vec x$ as $\no$. $\BO$ does not require an explicit representation of $f$, and instead search for near-optimal points is based on sampling the values of $f$ at a limited number of input points. $\BO$ is therefore used mainly for optimizing functions whose evaluation is expensive. In particular, $\BO$ is often used for hyper-parameter optimization when training $\ML$ models~\cite{Sno12}. During the last few decades, $\BO$ has been used extensively for designing engineering systems~\cite{Moc89}.

\BO iteratively builds a statistical model of $f(x)$ usually from a prior distribution defined by a Gaussian process~\cite{RW06}. At each iteration $i$ the current model is used to select the most promising candidate point $x_i$ to evaluate the objective function $f(x_i)$. This evaluation is used to update the posterior belief of the model. 
This process is repeated until some bound on the number of iterations is reached. 
 
 \BO behaves well in practice and usually can find $\no$ points in just a few iterations. 
 Nevertheless \BO has its limitations.
 In particular, since \BO is based on statistical approximations, there are no formal guarantees that the result achieved after
 a finite number of iterations 
 is actually optimal or even close to optimal.
 Another limitation of \BO is that even if the found point is near optimal the solution may not be stable: there may be close points on which the value of the objective function is very different from the optimum. 
 Such solutions are undesirable in many applications which require regions around solutions to be also near optimal. This motivates  combination \BO with SMT-based constraint solving to achieve both safe and stable solutions which we develop in this paper.

In this work we are interested in applying optimization techniques to optimize microprocessor design, in particular, analog components that can be modelled as real-valued functions.
In this context, the assumption that the objective function is very expensive to sample is no longer valid. Instead, achieving tighter approximations to the maximum  $\max f$ becomes more important even if computing $\no$ points that further improve the approximation accuracy becomes significantly more expensive.  Besides, in this context, it is important to guarantee that the computed $\no$ point $\vec x$ satisfies the design constraints and it is robust~\cite{BEYER20073190} in the sense that small perturbations to $\vec x$ still yield legal near-optimal points for the objective function~$f$. 

\begin{comment}
\strut\todofb{Is this paragraph relevant for the paper?}%
We use neural networks trained on input-output samples to approximate the design.
It is possible to further refine the training set at the input points of interest to improve the model's accuracy at the points where it matters the most, e.g., by following the \emph{proof based} model refinement procedure outlined in~\cite{DBLP:conf/fmcad/BrausseKK20}.  We can use this model as a representation of the objective function and from the $\no$ points
computed on this model, estimate using standard $\ML$ techniques the optimum of the original objective function. A detailed discussion of such an iterative model refinement procedure and how it tightens the probabilistic estimates on the optimum of the original objective function goes of beyond this work. Instead, here we focus on $\BO$ of functions specified explicitly rather than as a black-box, in particular, as a $\ML$ model.
So we can not only sample the objective function $f$ but we can use its explicit representation in the search for a maximum. We use SMT solvers to analyze the explicit representation. This is a significant paradigm shift compared to the widely accepted belief that $\BO$ is best suited for optimization of black-box functions.  In the experiments we use functions represented using neural networks with ReLU activation functions. It will be clear below that our approach is not limited to a neural network representation of $f$ and any representation for which efficient constraint solvers exist can be used.
\end{comment}

We propose a $\BO$ algorithm with the following properties:
\begin{itemize}
    \item \emph{Safety:} The computed $\no$ points are feasible (satisfy design constraints). In $\BO$ the feasibility of computed $\no$ solution is
    achieved by sampling the constraints~\cite{GKXWC14}. In contrast, in our approach we use the explicit representation of the constraints to guide the search for a feasible $\no$ point.
    
    \item \emph{Stability:} The computed $\no$ points are \emph{stable} in the sense that
     perturbations
    within user-specified regions preserve output within
    near-optimal value. 
    Stability is a critical requirement from analog devices because the inputs and the output can be perturbed due to uncertainties in the environment such as uncertain operating conditions, design parameter tolerances or actuator imprecisions~\cite{BEYER20073190}. This concept is studied in the $\BO$ setting as robustness~\cite{BSJC18,SEFR19,CKLP19},
    where it can be estimated with high confidence but cannot be proven formally.
    
    \item \emph{Accuracy:} Our algorithm can find safe and stable $\no$ points for which the value of the objective function $f$ is within a predefined distance from the real maximum, $\max f$. This is enabled by the fact that in our problem setting the objective function and the constraints are given explicitly: probabilistic methods cannot provide full guarantees due to a limited number of sampling of~$f$.   
\end{itemize}

The paper is organized as follows. In the next section we introduce the notation.
In Section~\ref{sec:bo} we recall the basics of $\BO$ and in Section~\ref{sec:stab} we define the stability and accuracy requirements from $\no$ solutions.  In Section~\ref{sec:opt}, we recall the algorithm for computing safe and stable configurations from~\cite{DBLP:conf/fmcad/BrausseKK20} and present a $\BO$ algorithm that combines the strengths of reasoning with probabilities (Bayesian inference) and constraint solving (SMT). Experimental results are presented in \Cref{sec:bench}. We conclude in \Cref{sec:concl}.

%%%%%%%%%%%%%%%%%%%%%%%%%%%%%%%%%%%%%%%%%%%%%%%%%%%%%%%%%%%%%%%%%%%%%%%%%%%%%%%
\section{Preliminaries}\label{sec:prelim}
%%%%%%%%%%%%%%%%%%%%%%%%%%%%%%%%%%%%%%%%%%%%%%%%%%%%%%%%%%%%%%%%%%%%%%%%%%%%%%%
Given a function $f:A\to B$, by $\dom f$ we denote its domain $A$.
Vectors $(x_1,\ldots,x_n)$ may occur in the abbreviated form $\vec x$.
Given $\vec a,\vec b\in\mathbb R^n$ with $a_i\leq b_i$ for all $i$,
by $[[\vec a,\vec b]]$ we denote the Cartesian product $\bigtimes_i[a_i,b_i]$ of their component-wise closed intervals $[a_i,b_i]$.
In this paper we consider formulas over $\langle\mathbb R,0,1,\mathcal F,P\rangle$,
where $P$ are the usual order predicates ${<},{\leq},{=}$, etc.\ and
$\mathcal F$ contains addition, multiplication with rational constants and
some non-linear functions for example $\operatorname{ReLU}: x\mapsto\max\{0,x\}$, that can be used to encode complex objective functions represented by neural networks.
Throughout, $x,y$ denote variables in formulas while $a,b,c,d,e,S,T$ stand for rational constants; both forms may be indexed.
Whenever we use a norm $\Vert\cdot\Vert$, we refer to the Chebyshev norm $(x_1,\ldots,x_n)\mapsto\max\{|x_1|,\ldots,|x_n|\}$.

\section{Bayesian optimization}\label{sec:bo}

A basic form of $\BO$ algorithm involves two primary components: a method for statistical inference, typically Gaussian process regression~\cite{RW06}; and an acquisition function for deciding where to sample next. The most popular choices for an acquisition function include expected improvement towards the optimum, knowledge gradient, and entropy search. \BO iteratively builds a statistical model of an objective function $f(\vec{x})$ from a prior distribution defined by a Gaussian process. At each iteration $i$ the current model is used to select the most promising candidate point $\vec x_i$ to evaluate the objective function $f(\vec{x}_i)$. This evaluation is used to update the posterior belief of the model. This process is repeated until bound $MaxIter$ on the number of iterations is reached. 

The basic optimisation algorithm for maximising a black-box function
$f:\mathbb R^n\to\mathbb R$, depicted in \Cref{alg:BO}, is based on a BO solver $A\Max$ with the following interface.
    \begin{itemize} %[\IEEEsetlabelwidth{\textsc{Observe}}]
    \item \textsc{Init}.
        Inputs: $\vec a,\vec b$ defining bounds $[[\vec a,\vec b]]\subseteq\dom f$, and a vector of initial points
        $(\vec x_i,y_i)_i$ satisfying $y_i=f(\vec x_i)$ for all $i$;
        Output: an initialised Bayesian optimiser
        for $f$ %$f:\mathbb R^n\to\mathbb R$
        with
        the posterior probability distribution on $f$ updated using the available samples $(\vec x_i,y_i)_i$.
    \item \textsc{Suggest}.
        Inputs: none.
        Output: A
        maximizer $\vec z\in[[\vec a,\vec b]]$
        of the acquisition function over $\vec x$, where the acquisition function is computed using the current posterior distribution.
    \item \textsc{Observe}.
        Inputs: $(\vec x,y)$.
        Output: the Bayesian optimiser with updated posterior probability based on $y = f(\vec{x})$.
    \end{itemize}

\begin{algorithm}
\caption{Basic optimisation algorithm using the \BO solver $A\Max$ for maximising $f(\vec x)$.}
\label{alg:BO}
\begin{algorithmic}
    \State $MaxIter$ -- a bound on the number of sampling iterations
    \State Sample $f$ at $n_0$ input points based on a heuristic:
    \State Let $\vec x_i\in [[\vec a,\vec b]]$ and $y_i=f(\vec x_i)$ for $i=1,\ldots,n_0$
    \State $A\gets A\Max.\textsc{Init}(\vec a,\vec b,(\vec x_i,y_i)_i)$
	\For {$n=n_0+1,\ldots,MaxIter$}
        \State $\vec x_n\gets A.\textsc{Suggest}$
		\State Sample $y_n = f(\vec{x}_n)$
		\State $A\gets A.\textsc{Observe}(\vec x_n,y_n)$
	\EndFor
	\State $y_j\gets\max\{y_1,\ldots,y_{MaxIter}\}$
	\State \Return a near-optimal solution: $(\vec x_j,y_j)$
\end{algorithmic}
\end{algorithm}

We will also use a minimising BO solver $B\Min$ that has the same interface.
This basic $\BO$ algorithm does not take as input any constraints on the inputs of $f$ thus it is not concerned with feasibility of the computed $\no$ point. Furthermore, this algorithm is not concerned with robustness of the computed $\no$ point either and does not give any estimates of how far from the real maximum the computed $\no$ solution is. We use this basic $\BO$ algorithm, complemented with
an SMT procedure where needed, as a building block of a more comprehensive, hybrid $\BO$ algorithm that gives formal guarantees that the computed $\no$ points are feasible, stable, with values
close to the real maximum up to arbitrary given accuracy.
We will define stability and accuracy
formally in the next section.

%%%%%%%%%%%%%%%%%%%%%%%%%%%%%%%%%%%%%%%%%%%%%%%%%%%%%%%%%%%%%%%%%%%%%%%%%%%%%%%
\section{Stability and Accuracy}\label{sec:stab}
%%%%%%%%%%%%%%%%%%%%%%%%%%%%%%%%%%%%%%%%%%%%%%%%%%%%%%%%%%%%%%%%%%%%%%%%%%%%%%%
Given a real-valued function $f$ on a bounded space $X\subset\mathbb R^n$,
we are interested in finding a maximum $f(\vec x^*)$ which is \emph{stable}, in particular, for all values  $\vec x'$ within a specified region around $\vec x^*$, $f(\vec x')$ stays within a specified threshold $\epsilon$ from $f(\vec x^*)$. This region around $\vec x^*$ is formalized as a \emph{stability guard}  $\theta(\vec x^*, \vec x')$ which we assume is a reflexive binary
relation over $\dom f$. This region could for example be defined by fixed radius
or one relative to $\Vert\vec x^*\Vert$, but could also take a more complicated shape and we do not impose any restrictions on it besides that of being encodable as a quantifier-free formula $\theta$.
We assume $f$ is encoded as the formula $F(x_1,\ldots,x_n,y)$ over variables $x_1,\ldots,x_n$ corresponding to the $n$ inputs and $y$ corresponding to the output $f(x_1,\ldots,x_n)$.

Given a stability guard $\theta$ for $f$ and a bounded subset $X$ of
$\dom f$, we are addressing the optimisation problem
\begin{equation}
    \max\nolimits_{\vec x}
    \mathop{\min\nolimits_{\vec x'}}\limits_{\theta(\vec x,\vec x')}
    f(\vec x')
    \label{eq:max-min}
\end{equation}
on $X$, and how to find reliable lower \emph{and} upper bounds on the optimum $y^*$, as well
as a corresponding $\vec x$.
That is, given a precision $\varepsilon>0$,
we want to compute $T$ and a point $\vec x\in X$
such that $T\leq y^*<T+\varepsilon$
and $T\leq f(\vec x')$ holds for all $\vec x'$ in the stability region around $\vec x$.
In particular, $T$ is an approximation of $y^*$ \emph{accurate} up to $\varepsilon$.
In the next section, we present algorithms based on combination of \BO and SMT that solve this problem.

%%%%%%%%%%%%%%%%%%%%%%%%%%%%%%%%%%%%%%%%%%%%%%%%%%%%%%%%%%%%%%%%%%%%%%%%%%%%%%%
\section{Optimisation procedure}\label{sec:opt}
%%%%%%%%%%%%%%%%%%%%%%%%%%%%%%%%%%%%%%%%%%%%%%%%%%%%%%%%%%%%%%%%%%%%%%%%%%%%%%%
The problem~\eqref{eq:max-min}
is equivalent to:
\begin{equation}
    \max y ~\text{s.t.}~
    \exists\vec x\,(\forall\vec x'\forall y'\,
    (\theta(\vec x,\vec x')\land F(\vec x',y')\to y\leq y'))
    \text.
    \label{eq:max}
\end{equation}
%where variables are restricted to their respective domains.
The type of formulas restricting $y$ in \eqref{eq:max} are also known to be in the \textsc{Gear}-fragment~\cite{DBLP:conf/fmcad/BrausseKK20} of $\exists^*\forall^*$ formulas.

We
first recall the decision procedure
\textsc{GearSAT$_\delta$}~\cite{DBLP:conf/fmcad/BrausseKK20} shown in \Cref{alg:1},
which we enhance with \BO solvers below.
It %\Cref{alg:1} 
 takes
a potential bound $T$ and either verifies $T$ to be a lower bound on the optimum $y^*$ or
proves it to be an upper bound on $y^*$.
\textsc{GearSAT$_\delta$} alternates two phases: search for candidate solution and search for counter-example for stability around the candidate solution. 
It does that by first finding a point $\vec x$ for which $f(\vec x)\geq T$ holds -- a \emph{candidate for a stable, accurate solution}. If there is none, clearly $T$ is an upper bound on $y^*$. Otherwise, it checks
whether when $\vec x$ is seen as the center of the stability region $\theta(\vec x,\cdot)$, there is a \emph{counter-example}
$\vec x'$, that is, $\theta(\vec x,\vec x')$ holds but $f(\vec x')\geq T$ does not (represented by $D_i(\vec x',y')$ constraint in \Cref{alg:1}).
In the case when there are no counter-examples with that property,
we can be sure that $\theta(\vec x,\vec x')$ implies
$f(\vec x')\geq T$ for all $\vec x'\in X$.
Otherwise, $\vec x'$ is a counter-example and the algorithm excludes the region $\theta(\cdot,\vec x')$ around it from the search for
the next candidate.
The stability condition $\theta$ guides the proof search by generating lemmas excluding regions around counter-examples.  

Here, $\delta\geq0$ refers to a constant which can be used to ensure that solutions have a distance of at least $\delta$ from regions defined by $\theta$ containing counter-examples to safety of $F$.
This is done by learning lemmas of the form $\neg\theta_\delta(\vec x,\vec d)$ for
each counter-example
$\vec d$, where $\theta_\delta$ is the $\delta$-relaxation of $\theta$ defined by
\[ \exists\vec z\,(\Vert\vec x-\vec z\Vert\leq\delta\land\theta(\vec z,\vec d))
   \text.
\]
Termination of \textsc{GearSAT$_\delta$} was shown in~\cite{DBLP:conf/fmcad/BrausseKK20}.
We can use \textsc{GearSAT$_\delta$} to find an optimal value with a precision $\epsilon$ on the value by a binary search of lower and upper bounds on the optimal value of the objective function: $T\leq y^*<T+\varepsilon$.

The key search parts in \textsc{GearSAT$_\delta$} rely on finding points (either candidates or counter-examples). 
One way to achieve this 
is by using
an SMT solver
to find points satisfying corresponding constraints (as done in~\cite{DBLP:conf/fmcad/BrausseKK20}) but this can be computationally expensive. In this work we propose to delegate the search part to \BO  and the certification part to SMT checks.
In this way we take the best from both worlds:  efficient search in complex spaces from \BO and formal guarantees on the optimisation results from SMT. 

Our algorithms do not depend on particular types of  \BO and SMT solvers used. We only assume that the SMT solver supports quantifier-free fragment including formulas $F$ and $\theta$.  Let us note that SMT solvers can handle a range of non-linear functions including polynomials, transcendental  functions such as combinations of sine, cosine, exponentials and solutions of differential equations~\cite{DBLP:conf/tacas/MouraB08,DBLP:conf/frocos/BrausseKKM19,DBLP:journals/corr/abs-2104-13269,DBLP:conf/tacas/CimattiGSS13}.

\begin{algorithm}
\caption{(\textsc{GearSAT}$_\delta$)
    General procedure for deciding whether a constant $T$ is a lower or upper bound on
    the solution $y^*$ of \eqref{eq:max}.
    }
\label{alg:1}
\begin{algorithmic}
    \State $F_1(\vec x,y)\gets F(\vec x,y)$
    \For {$i=1,2,3,\ldots$}
        \State $C_i(\vec x,y)\gets F_i(\vec x_i,y)\land y\geq T$
    	\State $\vec c_i\gets$ asgn.\ to $\vec x$ solution of $C_i(\vec x,y)$
    	    or \Return {\small\textsf{upper}}
    	%\State \Comment $\vec c_i$ is a candidate
    	\State $D_i(\vec x',y')\gets\theta(\vec c_i,\vec x')\land F_i(\vec x',y')\land y'<T$
    	\State $\vec d_i\gets$ asgn.\ to $\vec x'$ solution of $D_i(\vec x',y')$
    	    or \Return {\small\textsf{lower}} %$\vec c_i$
    	%\State \Comment $\vec d_i$ is a counter-example
    	\State $F_{i+1}(\vec x,y)\gets F_i(\vec x,y)\land\neg\theta_\delta(\vec x,\vec d_i)$
    \EndFor
\end{algorithmic}
\end{algorithm}

First we integrate \BO into  \textsc{GearSAT$_\delta$} for searching counter-examples.
We note that whenever $\theta_\delta(\vec c_i,\vec x')$ is equivalent to membership in the Cartesian product $[[\vec a_i,\vec b_i]]$ (as is always the case in our application),
Bayesian optimisation lends itself well to implement
the search for a counter-example by 
solving $D_i(\vec x',y)$, as is shown in
\Cref{alg:2}. 
This works by starting a new minimising \BO search in $[[\vec a_i,\vec b_i]]$
once \Cref{alg:1} found a candidate $\vec c_i$.

\begin{algorithm}
\caption{Finding counter-examples: Solving $D_i(\vec x',y')$ with Bayesian optimisation and SMT in the $i$-th iteration of \Cref{alg:1}.}
\label{alg:2}
\begin{algorithmic}
    \State Let $(\vec a_i,\vec b_i)$ denote the bounds on $\vec x'$ implied by $\theta(\vec c_i,\vec x')$ in $D_i(\vec x',y')$
    and let $T$ denote the bound on $y'$ in $D_i(\vec x',y')$
    \State Let $\vec e_1,\ldots,\vec e_k\in [[\vec a_i,\vec b_i]]$ \Comment \BO
    \If {$f(\vec e_j) < T$ for some $j\in\{1,\ldots,k\}$}
        \State $\vec d_i\gets\vec e_j$
    \Else
    \State $B_i\gets B\Min.\textsc{Init}(\vec a_i,\vec b_i,(\vec e_j,f(\vec e_j))_j)$
	\For {$j=1,\ldots,MaxIter$}
		\State ${\vec d_i}\gets B_i.\textsc{Suggest}$
		\State $B_i\gets B_i.\textsc{Observe}({\vec d_i},f({\vec d_i}))$
    	\IIf {$f({\vec d_i}) < T$}
    	    \Return %${\vec d_i}$
    	\EndIIf
	\EndFor
\State $\vec d_i\gets$ sol.\ of $D_i(\vec x',y')$ restricted to $\vec x'$ or \UNSAT \Comment\!SMT
	\EndIf
\end{algorithmic}
\end{algorithm}

Next, we integrate \BO for finding candidate solutions. 
For this we need to guide \BO to generate points outside of regions excluded by generated lemmas. 
Even though
most \BO solvers do not support constraints
like the lemmas $\neg\theta_\delta(\vec x,\vec d_m)$ learnt by \Cref{alg:1},
when $X\subseteq\dom f$ is
equivalent to membership in $[[\vec a,\vec b]]$ for some $\vec a,\vec b$,
it is still possible to use them in the search for a candidate $\vec c$
in our setting.
We achieve this by penalising any suggestion $\vec c$ that satisfies $\theta_\delta(\vec x,\vec d_m)$, for any counter-example $d_m$,  with the minimal value of a counter-example found when generating the lemma.
This is shown in \Cref{alg:3}.
Alongside the SMT solver it maintains the maximising BO solver $A_i$
that is initialised by
$A_1\gets A\Max.\textsc{Init}(\vec a,\vec b,(\vec e_j,f(\vec e_j))_j)$
for selected points $\vec e_j\in X$ which are either generated in previous iterations or randomly.

\begin{algorithm}
\caption{Finding candidates: Solving $C_i(\vec x,y)$ with Bayesian optimisation and SMT in the $i$-th iteration of \Cref{alg:1}. $A_i$ maximises.}
\label{alg:3}
\begin{algorithmic}
    \If {$i>1$} \Comment record previous counter-example
        \State $A_i\gets A_{i-1}.\textsc{Observe}(\vec c_{i-1},f(\vec d_{i-1}))$
    \EndIf
   	\For {$j=1,\ldots,MaxIter$}
   	\Comment \BO
        \State ${\vec c}_i\gets A_i.\textsc{Suggest}$
        \If {$\theta({\vec c}_i,\vec d_m)$ for any $m\in\{1,\ldots,i-1\}$} 
            \State $z\gets f(\vec d_m)$
            \Comment ${\vec c}_i$ is excluded by lemmas
        \Else
            \State $z\gets f({\vec c}_i)$
            \Comment ${\vec c}_i$ is an eligible candidate
        \EndIf
        \IIf {$z\geq T$}
            \Return %$\hat{\vec c}_i$ 
        \EndIIf
        \Comment $f(\vec c_i)\geq T$
        \State $A_i\gets A_i.\textsc{Observe}({\vec c}_i,z)$ \Comment $z<T$
    \EndFor
     \State $\vec c_i\gets$ solution of $C_i(\vec x,y)$ restricted to $\vec x$ or \UNSAT
     \Comment SMT
\end{algorithmic}
\end{algorithm}

Note that
due to penalising any of $A_i$'s suggestions inside regions
containing counter-examples
the function $A_i$ observes is not necessarily $f$.
Whenever $A_i$'s suggestion $\vec c$ lies in
a region around a previous counter-example $\vec d_m$ for some $m<i$,
that is, $\theta(\vec c,\vec d_m)$ holds,
we make $A_i$ believe that
the value of the function it optimises is $f(\vec d_m)$ instead of $f(\vec c)$.
Since $\vec d_m$ was a counter-example to safety of $F$,
specifically $f(\vec d_m)<T$, this has the effect of
penalising the suggestion $\vec c$
since it has already been proved to be $\theta$-close to a counter-example.
We call \textsc{GearSAT$_\delta$} with integrated  \BO for counter-examples and candidate search \textsc{GearSAT$_\delta$-\BO}.
As shown next, this exchange of
candidates and counter-examples
between BO and SMT solvers
proved to be
beneficial
in our experiments.

%%%%%%%%%%%%%%%%%%%%%%%%%%%%%%%%%%%%%%%%%%%%%%%%%%%%%%%%%%%%%%%%%%%%%%%%%%%%%%%
\section{Benchmarks}\label{sec:bench}
%%%%%%%%%%%%%%%%%%%%%%%%%%%%%%%%%%%%%%%%%%%%%%%%%%%%%%%%%%%%%%%%%%%%%%%%%%%%%%%
\textsc{GearSAT$_\delta$-\BO} is implemented in
our solver called \textsc{SMLP}.
As Bayesian optimizers $A\Max,B\Min$, we used the ones in the
\textsc{skopt} library~\cite{scikit-learn}
based on Gaussian processes,
with acquisition function \texttt{gp\_hedge}.
The SMT part is implemented using the state of the art solver
\textsc{Z3}~\cite{DBLP:conf/tacas/MouraB08}.

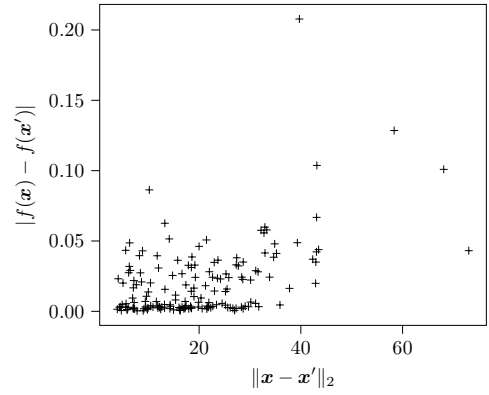
\begin{figure}
\centering
% This file was created by tikzplotlib v0.9.8.
\begin{tikzpicture}[scale=.75]

\begin{axis}[
tick align=outside,
tick pos=left,
x grid style={white!69.0196078431373!black},
xlabel={\(\displaystyle \Vert\vec x-\vec x'\Vert_2\)},
xmin=0.449251685986619, xmax=76.4725573071052,
xtick style={color=black},
y grid style={white!69.0196078431373!black},
ylabel={\(\displaystyle |f(\vec x)-f(\vec x')|\)},
ymin=-0.00998007085613784, ymax=0.218151052258252,
ytick style={color=black},
scatter/use mapped color={draw=black,fill=black},
    % <https://tex.stackexchange.com/a/96348>
    y tick label style={
        /pgf/number format/.cd,
            fixed,
            fixed zerofill,
            precision=2,
        /tikz/.cd
    },
]
\addplot [draw=black, fill=black, mark=+, only marks, scatter]
table{%
x  y
39.7102245207181 0.207781455753053
58.3446971263027 0.128515515150422
68.0900723735164 0.100898527714038
73.0169525061453 0.0431209771508143
10.2071854319659 0.0863278236412799
10.4319814785107 0.0202179345082363
19.1606789303652 0.032881302679248
27.390922239868 0.0380821515508252
17.249245334476 0.00702699165010623
25.1737147801187 0.0141036754708384
27.293741855501 0.0330709741843733
12.0153738154911 0.0308143463875472
17.885777054107 0.0327330456141643
42.9011623744077 0.0199071565650906
26.1112690453156 0.00254981396283371
31.1320872850974 0.028907912565715
23.0182512966813 0.0144394492277859
15.7887885006404 0.0362960758519393
14.1271674890356 0.0514835758211909
30.1204593843133 0.0222527073840113
35.8948172830706 0.00453086707032502
14.2583360936538 0.00192964171351262
5.63729735265723 0.00544723179155593
15.3657375407503 0.00810573501395506
21.7376032741637 0.00283965127117769
28.5848223648731 0.0226773541082774
6.91742504196029 0.00945152650838099
32.7467187933787 0.0557772926569657
16.6295239625079 0.0267388019169652
16.2942020602301 0.000645472064010555
8.45005698735029 0.0273654869039726
21.5464058813232 0.00377034004365684
6.32865936772373 0.0486504849347229
16.254120296681 0.001765252405594
31.7219751515318 0.00332509261128622
19.9861345154362 0.0461747513216664
8.99279579547057 0.000400198437429777
8.66402135194117 0.0209089624445855
28.6511853083221 0.00189064970213826
23.4196394825542 0.00470292677383655
30.1514271626945 0.00658646378017669
25.8309418034529 0.0239918391237945
9.64061769303103 0.0106997924860601
24.1938422459294 0.0229951325201794
8.26022676365575 0.0395751857804747
29.0486140378292 0.00365117302626317
18.5850023528806 0.0386565143371045
23.7029928476095 0.0363787064209316
33.8490036922634 0.0243194227160743
28.3663665889424 0.0242682821690603
19.2593794607769 0.0241705977979672
34.6531321307215 0.0384514757881071
28.6648311328628 0.0350445295346942
18.472033903614 0.031186152277614
43.0961225663248 0.0667809493120282
18.9715640135209 0.016499397619995
32.1723824690958 0.0576174386609642
22.6463572687192 0.0242463027262811
42.9480384900417 0.0350046544625648
16.642958583729 0.00307471834400941
11.5818691173782 0.00694525696457027
4.08978552393478 0.0231617775523454
29.6978057833538 0.00329849651674752
12.6638974121787 0.00305465394087601
5.03266430677648 0.0201164896203827
4.33163316016258 0.00289637411914612
9.15087306076375 0.00184735094570776
15.3645238771697 0.011522508033583
14.2799644596483 0.00462509249226017
18.4641182978528 0.0142407292658571
25.4856727782632 0.0158618872710183
23.5968518211179 0.0235189625193659
12.1728827020986 0.00315959033243196
22.2911477895594 0.0025461977193284
11.8493791629558 0.00397719542419872
17.3094101062381 0.00279210692663789
21.792357718958 0.0031341663465938
19.9047045501333 0.0063873794460485
9.78226155697614 0.0019065860409162
18.5018838646284 0.00216626268629017
11.9009267979977 0.00244909492534573
16.9916550601555 0.00181225320438871
34.8570613566467 0.0479813953068471
31.5814266785573 0.0281125453576299
6.14377029691787 0.027461416500826
35.2030049028723 0.0411570623700694
22.0111064301619 0.0281506244358569
17.1917346883268 0.0021356604213052
37.7689456453818 0.016326796101316
43.1468362941588 0.103726211949531
6.26787961158463 0.0319305865669415
9.95167314917115 0.00239038439425365
4.86634896405099 0.00471772597105158
24.5132876123004 0.00563515852080809
21.445302550291 0.0507430719578317
43.0784578069204 0.0422719439320862
5.86874014935687 0.00128524920523376
17.9609081128266 0.00185221477204056
20.3623631257476 0.0095077664896106
5.82444882266277 0.00116740330992871
20.5892116145718 0.00199325711017795
26.6342349119892 0.0020601509259901
7.20984295125552 0.00625398756591222
16.0409255305761 0.00271469564015225
9.47191260305478 0.00312572698093527
7.80292286543098 0.000389525649061806
3.90485648694656 0.00148102207127798
4.69877162549393 0.000560705460014299
9.41395336550082 0.00703747718167724
22.686320593701 0.00356919760626839
10.9229334825436 0.00302042720116025
6.45230136492515 0.0291032441112784
13.1899643259898 0.0048475428367587
32.949940675498 0.0415020173698415
21.9088349667567 0.00623818116651909
14.3593472541203 0.00360498396155684
7.07074320266288 0.00223162169963054
13.2141514801744 0.00238989690633562
9.8380319453268 0.00122589684046148
7.02840716216514 0.00223811824018072
22.9675086911033 0.0347837032024911
5.55967267963611 0.0433835692020856
27.7006052105276 0.0322681342428617
8.86054352475595 0.043017516978201
14.7818074472663 0.0254980947321892
11.750118048153 0.0395081999768094
42.3372415031638 0.0370315541825723
33.2921064824362 0.0578655477797807
7.17143700756416 0.0218471787027461
39.3144949118055 0.0487794575624478
17.4398795472525 0.00397627711479775
16.1913952551881 0.0006900319081935
28.3104475366182 0.00299859043027273
12.4407400022062 0.00146351803625544
21.5073797265523 0.00170035665383295
32.921107541276 0.0598217776843444
10.0053110893311 0.0137702374877864
7.04399021777896 0.00183700884967997
5.56002607657312 0.00286377903513446
19.8216097731045 0.00308575318688087
18.3840384772395 0.00267301978624923
7.31004767087578 0.00169908767307847
13.2534991168405 0.0626015521322276
10.4745286044324 0.0040176912097063
4.58477161166121 0.00325649424153207
13.2763331781275 0.0156561522119505
18.3230754733487 0.00365225809538527
25.6215180883473 0.00273467325727461
6.05218476005201 0.00290284343996006
14.9726673364599 0.00103414559805692
26.9722357623775 0.000668754657508153
12.2337366645138 0.00281736463785975
7.65154761485899 0.0188701709769248
31.1012784841748 0.00561904083340714
17.3964975099506 0.0188006740529403
21.2555649496246 0.0182105706719127
25.2869525950204 0.0265887200229193
25.1620494805956 0.00439361463107302
18.9831049453274 0.0104960934742828
7.06976734210825 0.0167115168096441
27.2656646147984 0.00207613417728214
43.4879126036346 0.0438669884781118
};
\end{axis}

\end{tikzpicture}
\caption{
    Correspondence between safety threshold
    and stability radius $\Vert\vec x-\vec x'\Vert_2$
    achievable on solutions found by a purely
    BO-based approach on instance
    3:1:1.
}\label{fig:1}
\end{figure}

We evaluated \textsc{GearSAT$_\delta$-\BO} (presented in \Cref{sec:opt}) on 6 industrial examples from the Electrical Validation Lab at Intel. 
These are neural network models representing signal integrity of transmitters and receivers  of a channel to a peripheral device.
This application requires solutions to be safe and stable (see, \Cref{sec:stab}), moreover the radii of stability regions are required to be proportional to the value of their respective centres.
We evaluated all 
4 combinations with and without {\BO}-guided searches for candidates
and counter-examples, respectively.
These results are shown in \Cref{tab:3}.
In the left-most column $i$ refers to the problem instance, $c$ to whether \BO search was used for candidates and $d$ to whether BO search was used for counter-examples.
Observations:
\begin{itemize}
\item
    Throughout our experiments, the combination of BO with SMT solvers
    proved to find the best bound $T=1$ to the optimum,
     whereas SMT alone timeouts in many cases.
\item
    This can be attributed to the facts that
    a) when BO failed to find counter-examples ($n_\mathrm{sa}$),
    none did exist ($N_\mathrm{sa}$), and that
    b) the average time taken to find a counter-example is %\todozk{$\approx39s$ vs. $0.7s$}
    much shorter for BO ($t_\mathrm{ce}/n_\mathrm{ce}$) than for
    SMT
    ($T_\mathrm{ce}/N_\mathrm{ce}$ for $N_\mathrm{ce}\neq0$ results). This suggests that BO constitutes a very good heuristic for finding counter-examples.
\item
    In addition, it turns out that
    as long as the total number of candidates found by BO-solver $A$ remains low ($n_\mathrm{cap}<50$), the asymptotically cubic complexity of the Gaussian process appears to be negligible compared to the overhead of rigorously solving the existential candidate formula by SMT.
    On the other hand, this advantage vanishes quickly after that,
    which motivates employing a kind of restart process similar to that successfully practised by current SAT and SMT solvers -- at least for the BO solver $A$.
    Initial experiments suggest that
    training samples for the restarted BO solver require careful selection.
\item
    The combinations *:0:1 correspond to those where BO tries to refute stability of
    SMT candidates. Throughout, it manages to do that with on average $n_\mathrm{cci}/n_\mathrm{ce}<3$ tries (iterations) in \Cref{alg:2}. On the other hand, for *:1:1 when BO tries to find counter-examples to BO candidates,
    this average with value $\approx8.8$ is much higher on average in our experiments.
   This suggests that the quality of BO candidates when guided by BO counter-examples (case *:1:1) compares favourably to that of the SMT candidates with BO counter-examples (case *:0:1).
\item
    All in all, we can see BO and SMT complement each other to solve
    the problem stated in \Cref{eq:max-min}. BO's
    ability to rapidly produce candidates and counter-examples
    initially  allows the combination to proceed to
    maximal regions quickly in most cases while still providing the formal
    guarantees on the validity and accuracy of stable solutions.
\item
\Cref{fig:1} shows the dependencies between the stability radius and the safety threshold achievable on candidate points. We can observe that initial candidates found by \BO are not necessarily stable, and therefore usage of SMT is essential to discover stable solutions.
\end{itemize}

\newcommand*\timeout{$>2$d}

\begin{table*}
\centering
\begin{tabular}{%
    @{\,}c@{:}c@{:}c@{~~}r@{~~~} % id th
    r@{ }r@{ }r@{ }r@{~~} % N_*
    r@{ }r@{ }r@{ }r@{~~~} % T_*
    r@{ }r@{ }r@{ }r@{ }r@{ }r@{ }r@{~~} % n_*
    r@{ }r@{ }r@{ }r@{~~} % t_*
    r@{ }%r@{ }r@{ }r@{ }r
}
%Inst & BO-ca & BO-ce
\multicolumn{3}{@{\,}l}{$i\text:c\text:d$}
& $T$ & N$_{\mathrm{cap}}$ & N$_{\mathrm{can}}$ & N$_{\mathrm{ce}}$ & N$_{\mathrm{sa}}$ & T$_{\mathrm{cap}}$ & T$_{\mathrm{can}}$ & T$_{\mathrm{ce}}$ & T$_{\mathrm{sa}}$ & n$_{\mathrm{cai}}$ & n$_{\mathrm{cci}}$ & n$_{\mathrm{cap}}$ & n$_{\mathrm{can}}$ & n$_{\mathrm{ce}}$ & n$_{\mathrm{sa}}$ & n$_{\mathrm{un}}$ & t$_{\mathrm{cap}}$ & t$_{\mathrm{can}}$ & t$_{\mathrm{ce}}$ & t$_{\mathrm{sa}}$ %& user & sys & cpu
    & time %wall %& mem
    \\ \hline
0 & 0 & 0 & $\geq0$.80 & 1376 & 0 & 1375 & 1 & 55114.7 & 0.0 & 54757.5 & 20.3 & 0 & 0 & 0 & 0 & 0 & 0 & 0 & 0.0 & 0.0 & 0.0 & 0.0 & \timeout \\
0 & 0 & 1 & $\geq0$.80 & 5049 & 0 & 0 & 1 & 105017.1 & 0.0 & 0.0 & 21.2 & 0 & 6711 & 0 & 0 & 5048 & 1 & 0 & 0.0 & 0.0 & 829.8 & 81.3 & \timeout \\
0 & 1 & 0 & $\geq0$.95 & 19 & 0 & 30 & 2 & 471.6 & 0.0 & 92.0 & 5.3 & 1114 & 0 & 13 & 19 & 0 & 0 & 0 & 379.1 & 105059.2 & 0.0 & 0.0 & \timeout \\
0 & 1 & 1 & $\geq0$.95 & 3 & 0 & 0 & 2 & 233.3 & 0.0 & 0.0 & 9.2 & 1113 & 1333 & 254 & 3 & 255 & 2 & 0 & 69931.5 & 19954.4 & 248.4 & 58.1 & \timeout \\
\hline
1 & 0 & 0 & $\geq0$.80 & 3254 & 0 & 3253 & 1 & 80374.5 & 0.0 & 30002.8 & 18.8 & 0 & 0 & 0 & 0 & 0 & 0 & 0 & 0.0 & 0.0 & 0.0 & 0.0 & \timeout \\
1 & 0 & 1 & $\geq0$.50 & 8094 & 0 & 0 & 0 & 34519.6 & 0.0 & 0.0 & 0.0 & 0 & 16674 & 0 & 0 & 8094 & 0 & 0 & 0.0 & 0.0 & 835.5 & 0.0 & \timeout \\
1 & 1 & 0 & $\geq1$.00 & 5 & 0 & 111 & 3 & 389.0 & 0.0 & 332.5 & 31.9 & 1163 & 0 & 109 & 5 & 0 & 0 & 0 & 61212.0 & 43901.3 & 0.0 & 0.0 & \timeout \\
1 & 1 & 1 & 1.00 & 2 & 0 & 0 & 4 & 322.2 & 0.0 & 0.0 & 29.2 & 525 & 423 & 74 & 2 & 72 & 4 & 0 & 6125.5 & 859.5 & 33.5 & 330.6 %& 7015 & 9117 & 193\%
    & 8353 %& 868
    \\
\hline
2 & 0 & 0 & $\geq0$.95 & 1025 & 0 & 1023 & 2 & 31240.7 & 0.0 & 84147.1 & 87.8 & 0 & 0 & 0 & 0 & 0 & 0 & 0 & 0.0 & 0.0 & 0.0 & 0.0 & \timeout \\
2 & 0 & 1 & $\geq0$.50 & 5000 & 0 & 0 & 0 & 36717.1 & 0.0 & 0.0 & 0.0 & 0 & 6420 & 0 & 0 & 5001 & 0 & 0 & 0.0 & 0.0 & 349.4 & 0.0 & \timeout \\
2 & 1 & 0 & $\geq1$.00 & 21 & 0 & 31 & 3 & 1165.4 & 0.0 & 195.1 & 27.1 & 1239 & 0 & 13 & 21 & 0 & 0 & 0 & 364.9 & 101490.2 & 0.0 & 0.0 & \timeout \\
2 & 1 & 1 & $\geq1$.00 & 12 & 0 & 0 & 3 & 1153.9 & 0.0 & 0.0 & 22.0 & 1247 & 359 & 60 & 12 & 69 & 3 & 0 & 16773.8 & 67275.1 & 64.4 & 180.3 & \timeout \\
\hline
3 & 0 & 0 & $\geq0$.50 & 2608 & 0 & 2608 & 0 & 70288.7 & 0.0 & 42255.8 & 0.0 & 0 & 0 & 0 & 0 & 0 & 0 & 0 & 0.0 & 0.0 & 0.0 & 0.0 & \timeout \\
3 & 0 & 1 & $\geq0$.95 & 9301 & 0 & 1154 & 2 & 23559.8 & 0.0 & 12381.9 & 82.9 & 0 & 21674 & 0 & 0 & 8145 & 2 & 1154 & 0.0 & 0.0 & 1135.9 & 111.4 & \timeout \\
3 & 1 & 0 & 1.00 & 0 & 0 & 16 & 4 & 0.0 & 0.0 & 93.7 & 33.9 & 110 & 0 & 20 & 0 & 0 & 0 & 0 & 201.3 & 0.0 & 0.0 & 0.0 %& 328 & 486 & 204\%
    & 399 %& 385
    \\
3 & 1 & 1 & 1.00 & 0 & 0 & 0 & 4 & 0.0 & 0.0 & 0.0 & 21.6 & 96 & 365 & 22 & 0 & 18 & 4 & 0 & 118.0 & 0.0 & 42.6 & 281.6 %& 475 & 1082 & 302\% 
    & 514 %& 377
    \\
\hline
4 & 0 & 0 & $\geq0$.80 & 1112 & 0 & 1110 & 1 & 8193.1 & 0.0 & 107511.3 & 37.0 & 0 & 0 & 0 & 0 & 0 & 0 & 0 & 0.0 & 0.0 & 0.0 & 0.0 & \timeout \\
4 & 0 & 1 & $\geq0$.80 & 1406 & 0 & 0 & 1 & 104095.2 & 0.0 & 0.0 & 44.3 & 0 & 3426 & 0 & 0 & 1405 & 1 & 0 & 0.0 & 0.0 & 3168.8 & 138.6 & \timeout \\
4 & 1 & 0 & $\geq0$.95 & 4 & 0 & 136 & 2 & 1022.2 & 0.0 & 498.7 & 12.4 & 1098 & 0 & 134 & 4 & 0 & 0 & 0 & 70635.3 & 27534.9 & 0.0 & 0.0 & \timeout \\
4 & 1 & 1 & $\geq0$.95 & 6 & 0 & 0 & 2 & 917.3 & 0.0 & 0.0 & 6.4 & 1143 & 221 & 64 & 6 & 68 & 2 & 0 & 48538.4 & 39695.9 & 10.1 & 118.4 & \timeout \\
\hline
5 & 0 & 0 & $\geq0$.50 & 1446 & 0 & 1446 & 0 & 42628.9 & 0.0 & 70583.7 & 0.0 & 0 & 0 & 0 & 0 & 0 & 0 & 0 & 0.0 & 0.0 & 0.0 & 0.0 & \timeout \\
5 & 0 & 1 & $\geq0$.50 & 14725 & 0 & 388 & 0 & 95689.8 & 0.0 & 1709.6 & 0.0 & 0 & 27543 & 0 & 0 & 14337 & 0 & 388 & 0.0 & 0.0 & 1077.3 & 0.0 & \timeout \\
5 & 1 & 0 & 1.00 & 0 & 0 & 29 & 4 & 0.0 & 0.0 & 131.8 & 27.5 & 209 & 0 & 33 & 0 & 0 & 0 & 0 & 852.7 & 0.0 & 0.0 & 0.0 %& 827 & 1275 & 175\%
    & 1195 %& 439
    \\
5 & 1 & 1 & 1.00 & 0 & 0 & 0 & 4 & 0.0 & 0.0 & 0.0 & 25.7 & 161 & 382 & 33 & 0 & 29 & 4 & 0 & 478.4 & 0.0 & 29.5 & 414.4 %& 695 & 1586 & 206\% 
    & 1107 %& 388
    \\
\end{tabular}
\caption{Indicators of SMLP for all combinations of en-/disabled BO heuristics for candidates and counter-examples. $N_*,n_*$: number of SMT/BO solutions. $T_*,t_*$: time of SMT/BO solutions.
$\mathrm{ca(p|n)}$: $\exists$ or $\neg\exists$ candidate.
$\mathrm{ce|sa}$: $\exists$ or $\neg\exists$ counter-example.
$\mathrm{cai|cci}$: total BO calls for candidate or counter-example.
$\mathrm{un}$: BO library failure.
}\label{tab:3}
\end{table*}

%%%%%%%%%%%%%%%%%%%%%%%%%%%%%%%%%%%%%%%%%%%%%%%%%%%%%%%%%%%%%%%%%%%%%%%%%%%%%%%
\section{Conclusions}\label{sec:concl}
%%%%%%%%%%%%%%%%%%%%%%%%%%%%%%%%%%%%%%%%%%%%%%%%%%%%%%%%%%%%%%%%%%%%%%%%%%%%%%%

We have introduced a hybrid optimization algorithm \textsc{GearSAT$_\delta$-\BO} that uses Bayesian optimization and SMT solvers as
its building blocks.
SMT solving is used to establish formal guarantees to optimality and stability
of the computed
solutions.
$\BO$ on the other hand
is used to suggest valuable candidates towards stable
$\no$ solutions and significantly speeds up the overall search.
In this way we combine the strengths of both approaches:
the power of 
statistical inference by BO to guide the search with formal guarantees provided by SMT.

To the best of our knowledge this is the first work that combines $\BO$ and SMT solving to overcome  basic limitations of the $\BO$, in particular, its inability to give formal guarantees 
 %(full proof) 
of stability and accuracy of the computed optimum, which becomes possible to resolve in cases when a representation of function is given explicitly rather as a black-box.

We believe that the observation that $\BO$ is
very good in finding counter-examples in large
multi-dimensional spaces, opens up new opportunities for applying $\BO$ for counter-example generation and directing the search in multiple areas of automated reasoning and formal verification. 

\paragraph*{Acknowledgement}
This research was supported by a grant from Intel Corporation.

%\FloatBarrier

\bibliographystyle{IEEEtran}
\bibliography{IEEEabrv,references}

\begin{thebibliography}{10}
\providecommand{\url}[1]{#1}
\csname url@samestyle\endcsname
\providecommand{\newblock}{\relax}
\providecommand{\bibinfo}[2]{#2}
\providecommand{\BIBentrySTDinterwordspacing}{\spaceskip=0pt\relax}
\providecommand{\BIBentryALTinterwordstretchfactor}{4}
\providecommand{\BIBentryALTinterwordspacing}{\spaceskip=\fontdimen2\font plus
\BIBentryALTinterwordstretchfactor\fontdimen3\font minus
  \fontdimen4\font\relax}
\providecommand{\BIBforeignlanguage}[2]{{%
\expandafter\ifx\csname l@#1\endcsname\relax
\typeout{** WARNING: IEEEtran.bst: No hyphenation pattern has been}%
\typeout{** loaded for the language `#1'. Using the pattern for}%
\typeout{** the default language instead.}%
\else
\language=\csname l@#1\endcsname
\fi
#2}}
\providecommand{\BIBdecl}{\relax}
\BIBdecl

\bibitem{Moc75}
J.~Mockus, ``On bayesian methods for seeking the extremum,'' \emph{Optimization
  Techniques IFIP Technical Conference}, 1975.

\bibitem{Fra18}
P.~I. Frazier, ``A tutorial on bayesian optimization,'' \emph{CoRR}, vol.
  abs/1807.02811, 2018.

\bibitem{Sno12}
J.~Snoek, H.~Larochelle, and R.~P. Adams, ``Practical bayesian optimization of
  machine learning algorithms,'' \emph{Advances in Neural Information
  Processing Systems}, pp. 2951--2959, 2012.

\bibitem{Moc89}
J.~Mockus, ``Bayesian approach to global optimization: Theory and
  applications,'' \emph{Kluwer Academic Publishers}, 1989.

\bibitem{RW06}
C.~E. Rasmussen and C.~K.~I. Williams, ``Gaussian processes for machine
  learning,'' \emph{The MIT Press, Cambridge, MA}, 2006.

\bibitem{BEYER20073190}
H.-G. Beyer and B.~Sendhoff, ``Robust optimization – a comprehensive
  survey,'' \emph{Computer Methods in Applied Mechanics and Engineering}, vol.
  196, no.~33, pp. 3190--3218, 2007.

\bibitem{GKXWC14}
J.~R. Gardner, M.~J. Kusner, Z.~E. Xu, K.~Q. Weinberger, and J.~P. Cunningham,
  ``Bayesian optimization with inequality constraints,'' in \emph{Proceedings
  of the 31th International Conference on Machine Learning, {ICML} 2014,
  Beijing, China}, ser. {JMLR} Workshop and Conference Proceedings,
  vol.~32.\hskip 1em plus 0.5em minus 0.4em\relax JMLR.org, 2014, pp. 937--945.

\bibitem{BSJC18}
I.~Bogunovic, J.~Scarlett, S.~Jegelka, and V.~Cevher, ``Adversarially robust
  optimization with gaussian processes,'' in \emph{Advances in Neural
  Information Processing Systems 31: NeurIPS 2018}, S.~Bengio, H.~M. Wallach,
  H.~Larochelle, K.~Grauman, N.~Cesa{-}Bianchi, and R.~Garnett, Eds., 2018, pp.
  5765--5775.

\bibitem{SEFR19}
N.~D. Sanders, R.~M. Everson, J.~E. Fieldsend, and A.~A.~M. Rahat., ``A
  {B}ayesian approach for the robust optimisation of expensive-to-evaluate
  functions,'' \emph{IEEE Transactions on Evolutionary Computing. arXiv :
  1904.11416v2}, pp. 2951--2959, 2019.

\bibitem{CKLP19}
L.~Cardelli, M.~Kwiatkowska, L.~Laurenti, and A.~Patane, ``Robustness
  guarantees for bayesian inference with gaussian processes,'' in \emph{The
  Thirty-Third {AAAI} Conference on Artificial Intelligence, {AAAI} 2019, The
  Thirty-First Innovative Applications of Artificial Intelligence Conference,
  {IAAI} 2019, The Ninth {AAAI} Symposium on Educational Advances in Artificial
  Intelligence, {EAAI} 2019, Honolulu, Hawaii, USA}.\hskip 1em plus 0.5em minus
  0.4em\relax {AAAI} Press, 2019, pp. 7759--7768.

\bibitem{DBLP:conf/fmcad/BrausseKK20}
F.~Brau{\ss}e, Z.~Khasidashvili, and K.~Korovin, ``Selecting stable safe
  configurations for systems modelled by neural networks with {ReLU}
  activation,'' in \emph{2020 Formal Methods in Computer Aided Design, {FMCAD}
  2020, Haifa, Israel}.\hskip 1em plus 0.5em minus 0.4em\relax {IEEE}, 2020,
  pp. 119--127.

\bibitem{DBLP:conf/tacas/MouraB08}
L.~M. de~Moura and N.~Bj{\o}rner, ``{Z3:} an efficient {SMT} solver,'' in
  \emph{Tools and Algorithms for the Construction and Analysis of Systems, 14th
  International Conference, {TACAS} 2008, Held as Part of the Joint European
  Conferences on Theory and Practice of Software, {ETAPS} 2008, Budapest,
  Hungary. Proceedings}, ser. LNCS, C.~R. Ramakrishnan and J.~Rehof, Eds., vol.
  4963.\hskip 1em plus 0.5em minus 0.4em\relax Springer, 2008, pp. 337--340.

\bibitem{DBLP:conf/frocos/BrausseKKM19}
F.~Brau{\ss}e, K.~Korovin, M.~V. Korovina, and N.~T. M{\"{u}}ller, ``A
  {CDCL}-style calculus for solving non-linear constraints,'' in
  \emph{Frontiers of Combining Systems - 12th International Symposium, FroCoS
  2019, London, UK. Proceedings}, ser. LNCS, A.~Herzig and A.~Popescu, Eds.,
  vol. 11715.\hskip 1em plus 0.5em minus 0.4em\relax Springer, 2019, pp.
  131--148.

\bibitem{DBLP:journals/corr/abs-2104-13269}
F.~Brau{\ss}e, K.~Korovin, M.~V. Korovina, and N.~T. M{\"u}ller, ``The ksmt
  calculus is a {\(\delta\)}-complete decision procedure for non-linear
  constraints,'' in \emph{The 28th International Conference on Automated
  Deduction, CADE-28 2021. Proceedings}, ser. LNCS, A.~Platzer and
  G.~Sutcliffe, Eds.\hskip 1em plus 0.5em minus 0.4em\relax Springer, 2021 (to
  appear).

\bibitem{DBLP:conf/tacas/CimattiGSS13}
A.~Cimatti, A.~Griggio, B.~J. Schaafsma, and R.~Sebastiani, ``The {MathSAT5}
  {SMT} solver,'' in \emph{Tools and Algorithms for the Construction and
  Analysis of Systems - 19th International Conference, {TACAS} 2013, Held as
  Part of the European Joint Conferences on Theory and Practice of Software,
  {ETAPS} 2013. Proceedings}, ser. LNCS, N.~Piterman and S.~A. Smolka, Eds.,
  vol. 7795.\hskip 1em plus 0.5em minus 0.4em\relax Springer, 2013, pp.
  93--107.

\bibitem{scikit-learn}
F.~Pedregosa, G.~Varoquaux, A.~Gramfort, V.~Michel, B.~Thirion, O.~Grisel,
  M.~Blondel, P.~Prettenhofer, R.~Weiss, V.~Dubourg, J.~Vanderplas, A.~Passos,
  D.~Cournapeau, M.~Brucher, M.~Perrot, and E.~Duchesnay, ``Scikit-learn:
  Machine learning in {P}ython,'' \emph{Journal of Machine Learning Research},
  vol.~12, pp. 2825--2830, 2011.

\end{thebibliography}

\end{document}